\documentclass[letterpaper, 10 pt, conference]{ieeeconf}
\IEEEoverridecommandlockouts

\usepackage{graphics}
\usepackage{mathtools}
\usepackage{xcolor}
\usepackage[caption=false, font=scriptsize, labelfont=sf, textfont=sf]{subfig}
\usepackage{amssymb}
\usepackage[ruled,vlined]{algorithm2e}
\usepackage[noadjust]{cite}

\title{\LARGE \bf
A hybrid model-based evolutionary optimization with passive boundaries for physical human-robot interaction}

\author{Gustavo J. G. Lahr$^1$, Henrique B. Garcia$^2$, Arash Ajoudani$^1$, Thiago Boaventura$^2$, and Glauco A. P. Caurin$^2$
\thanks{$^{1}$Gustavo J. G. Lahr, and Arash Ajoudani are with the Human-Robot Interfaces and Physical Interaction Lab (HRI2), Istituto Italiano di Tecnologia, 16163 Genoa, Italy. Email: {\tt\small gustavo.giardini@iit.it}}
\thanks{$^{2}$Henrique B. Garcia, Thiago Boaventura, and Glauco A. P. Caurin are with the São Carlos School of Engineering, University of São Paulo, São Carlos, Brazil.}
}

\begin{document}

\maketitle
\thispagestyle{empty}
\pagestyle{empty}

\begin{abstract}

The field of physical human-robot interaction has dramatically evolved in the last decades. As a result, the robotic system's requirements have become more challenging, including personalized behavior for different tasks and users. Various machine learning techniques have been proposed to give the robot such adaptability features.
This paper proposes a model-based evolutionary optimization algorithm to tune the apparent impedance of a wrist rehabilitation device.
We used passivity to define boundaries for the possible controller outcomes, limiting the shared autonomy of the robot and ensuring the coupled system stability.
The experiment consists of a hardware-in-the-loop optimization and a one-degree-of-freedom robot used for wrist rehabilitation. Experimental results showed that the proposed technique could generate customized passive impedance controllers for three subjects. Furthermore, when compared with a constant impedance controller, the method suggested decreased in 20\% the root mean square of interaction torques while maintaining stability during optimization.
\end{abstract}

\section{Introduction}

Being capable of handling physical interaction is desirable and mandatory in a diverse and growing number of robotic applications, including exoskeletons, robotic surgery, industrial assembly tasks, and collaborative robotics.
In particular, the field of physical Human-Robot Interaction (pHRI) has made significant progress in the last decades \cite{haddadin2016physical},
providing, e.g., personalized behavior in collaborative, assistive, and rehabilitation tasks \cite{krebs1998robot,ajoudani2018progress,kubota2020jessie}.

Such modern pHRI features use machine learning and optimization techniques to enhance the robot's capability of co-adapting to a specific user and task \cite{ikemoto2012physical,peternel2016adaptation,tsiakas2017interactive}, including changes in the robot's apparent impedance \cite{ficuciello2015variable,li2017adaptive,li2018stable}.
Although such techniques aim to give the robot a certain degree of autonomy to change its dynamic behavior, the shared autonomy must be limited to ensure the coupled system is safe, reducing the risk of potential injuries \cite{pervez2008safe}.

\begin{figure}
    \centering
    \includegraphics[width=.52\linewidth, trim={1.0cm 2.5cm 1.0cm 7cm},clip]{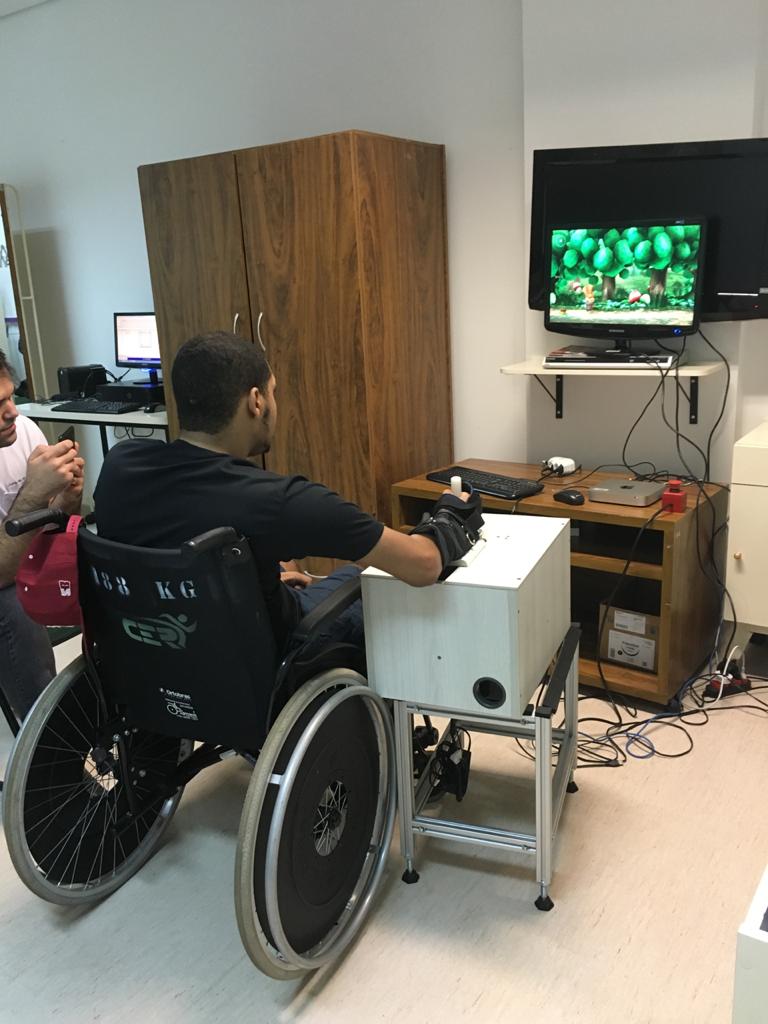}
    \caption{Robot for wrist rehabilitation therapy with serious games \cite{Andrade2018}. Each patient has different muscle stiffness and pain sensitivity, so the robot must adapt for better treatment.
    }
    \label{fig:case}
\end{figure}

Evolutionary optimization presents several interesting capabilities for adapting the robot behavior in pHRI tasks. Since it is hard to obtain an accurate model for each patient beforehand, it becomes desirable to have techniques capable of dealing with unknown environments during the interaction, which is the case of evolutionary optimization. These methods are motivated by natural processes and run on population-based metaheuristics to approximate the optimal results in problems that are hard to solve by other techniques. First, they do not use an explicit objective function, implying that neither the robot nor environment models are needed \cite{Lahr2017}. Second, evolutionary algorithms are less susceptible to local minima due to statistical population sampling \cite{Deb2002}. Third, they can decouple the metrics for multiobjective optimization instead of creating new objective functions that depend on other hyper-parameters' choices \cite{Hogan2020}.

The main challenge of evolutionary optimization is the slow convergence due to the many trials the system must run, especially when working with hardware-in-the-loop optimization \cite{Nadeau2018, Lahr2020}. Furthermore, an unconstrained optimization may employ parameters that could lead to instability, especially when dealing with humans.

A solution to accelerate convergence and avoid unstable scenarios is to narrow the search space by specifying boundaries on the set of possible controllers.
In this paper, we propose to add passive constraints to an evolutionary optimization algorithm, specifically a genetic algorithm, to enhance the control stability and adaptation when the robot is coupled to a human user.
The concept of passivity is very often used to guarantee the stability of coupled systems during physical interaction \cite{colgate94}, including when humans are involved \cite{buerger2006relaxing}.
By restricting the search to the range of stiffness and damping that can be passively emulated by the robot impedance controller, named \emph{Z-width} \cite{colgate1994factors}, we also accelerate the convergence.

The main contributions of this paper are twofold: 1) faster optimization convergence using a hybrid evolutionary and model-based technique with passive constraints, and 2) improved robot autonomy during pHRI by allowing continuous optimization during operation with stable setups.

\section{Related work}

Evolutionary techniques have been widely used in robotics. Applications range from humanoids to topological design \cite{Chang2020, RadhakrishnaPrabhu2018surveyEvolutionary}. Optimizing physical interaction with the robot's environment is also a topic of interest in the literature, especially for a suitable impedance controller for each task. Li \cite{Li2016} optimized an impedance controller along one degree-of-freedom (DOF) in a simulated environment without considering the transition from free motion to constrained motion. Lahr et al. \cite{Lahr2017} optimized an admittance controller for industrial robots along one DOF in an experimental setup, optimizing in an experimental design taking into account several nonlinearities.
Nadeau and Bonev \cite{Nadeau2018} used an evolutionary algorithm to optimize the impedance controller during a human-machine task with multiple DOFs using an industrial robot. In contrast, Lahr et al. \cite{Lahr2020} optimized an admittance controller for an industrial robot with three-DOF for an assembly task using dimensionality reduction for faster convergence.
Mehdi and Boubaker \cite{mehdi2011impedance} used Particle Swarm Optimization (PSO) algorithm to optimize an impedance controller for a simulated planar three-DOF manipulator.

The above contributions did not add equality or inequality constraints to the search space. Instead, they just defined the limits that the optimization should be conducted. Moreover, these search space limits are defined empirically, which is usually very time-consuming and conservative, as it may exclude sets of controllers capable of leading to stable behavior and improved performance.

Other studies used a model-based approach to design interaction controllers. For example, Averta and Hogan \cite{Hogan2020} optimized passive controllers in a multiobjective problem using the classical reduction to single-objective optimization, focusing on one optimal solution.
Colgate and Brown \cite{colgate1994factors} studied the impact of factors on the Z-width of haptic devices.
Boaventura et al. \cite{Boaventura2013} showed the influence of practical hydraulics actuation aspects, such as valve bandwidth, on the Z-width of a legged robot.
Calanca et al. \cite{Calanca2017} compared the experimental and theoretical Z-width of several interaction controllers for Series Elastic Actuators (SEAs) along one DOF.
Also, Collonese and Okamura \cite{Colonnese2015} presented the set of virtual masses capable of being rendered by the robot, named \emph{M-width}. However, none of these model-based papers optimized the interaction for best performance.

\section{Impedance optimization}
\label{sec:impOpt}

\subsection{Interaction controller}

Impedance control is widely used for interaction tasks. This controller regulates the power exchange between the robot and the environment or user, thus dealing with the relationship between generalized force and velocity at the interaction port \cite{Hogan1985}. Along one rotational DOF, the relationship between torque $\tau$ and angular velocity $\dot{q}$ defines the impedance $Z(s)=\tau(s)/\dot{q}(s)$. For a linear system, the inverse of the impedance is called admittance $Y(s) = \dot{q}(s)/\tau(s)$.

\begin{figure}[t]
	\centering
  	\includegraphics[width=.85\linewidth]{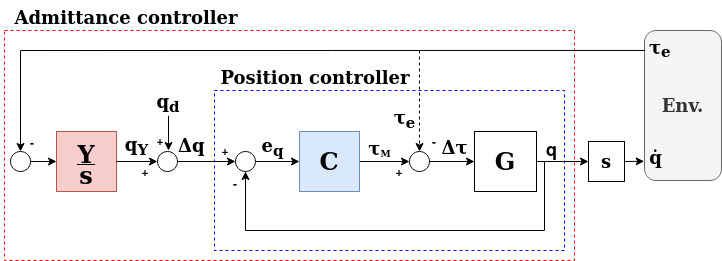}
    \caption{
    Block diagram for the complete admittance controller. The inner loop controls position $q$ for the desired position $q^d$, where $e_q$ is the position error, and $\tau_M$ is the torque generated by the motor. $\Delta \tau$ is the net torque from external torque $\tau_e$ (from environment interaction) and motor's torque, working as input for the plant $G$. External torques go through the admittance controller $Y$, generating a correction $q_Y$ to the desired position. $\Delta q$ is the actual position tracked by the position controller.}
    \label{fig:block_diagram_admit}
\end{figure}

The system considered in this work is a one-DOF platform used for human-robot interaction \cite{Andrade2018}, shown in Fig. \ref{fig:case} and Fig. \ref{fig:experimental-setup}, composed of a DC brushed motor, a gear transmission, and a handler for gripping. Fig. \ref{fig:block_diagram_admit} shows the complete block diagram, where the handler $G(s)$ has its position $q$ controlled in the internal loop by the position controller $C(s)$. At the same time, external torques $\tau_e$ may be applied directly to the motor (the dashed line represents natural feedback). Torques are also fed back via a force-torque sensor to the admittance controller $Y(s)$ through the outer loop. Both inputs have their impact at the final position shown by the following transfer function:

\begin{equation}
\begin{aligned}
    q(s) = &\frac{G(s)C(s)}{1+G(s)C(s)}q^d + \\ &\frac{G(s)(1+C(s)Y(s))}{1+G(s)C(s)}(-\tau_{e}) \
\end{aligned}
    \label{eq:tf_q}
\end{equation}

The transfer functions are: 

\begin{itemize}
    \item Handler: $G(s)=\frac{1}{Js^2}$
    \item Admittance controller: $\frac{Y}{s}(s) = \frac{s}{B_y s+ K_y}$
    \item Position controller: $C(s)=(Ds+P)k_r$
\end{itemize}

\noindent with $J$ being the handler's inertia, and $B_y$ and $K_y$ the admittance's controller damping and stiffness, respectively. Considering the position controller, $P$ is  its proportional gain, and $D$, the derivative. A gear ratio of $k_r$ is coupled between the motor and the handler. The friction of the gear is neglected in the modeling. All gains and parameters are strictly positive. By setting the desired position $q^d$ to zero, it is possible to write the system's resulting impedance as follows:

\begin{equation}
    Z(s)=\frac{-\tau_{e}(s)}{\dot{q}(s)} = \frac{[Js^2+(Ds+P)k_r](B_y s+K_y)}{s[Dk_rs^2+(B_y + Pk_r)s+K_y]} \ .
    \label{eq:impedance_siso}
\end{equation}

\subsection{Passivity}

Passivity is a property of a dynamical system that does not increase the externally supplied energy to the same system. This property is mainly influential during the design of interaction controllers because the interaction between two passive systems results in a stable coupled system. On the other hand, the interaction of two stable but not passive systems may result in unstable coupled systems \cite{Brogliato2007}.

Mathematically, passivity is defined as the relationship between a dynamic system's input $u(t)$ and its output $y(t)$, where the output is assumed to be dependent on the input. Thus, a system is said to be passive from $u(t)$ to $y(t)$ if it exists a constant $\beta \geq 0$ such that

\begin{equation}
    \int^{\tau}_{0}y(t)u(t)dt \geq -\beta \ .
    \label{eq:passivity}
\end{equation}

As the impedance transfer function given by (\ref{eq:impedance_siso}) is linear and time-invariant, passivity is achieved if the following two necessary conditions are met \cite{Boaventura2013, Calanca2017}:

\begin{enumerate}
    \item $Z(s)$ has no poles in the right half-plane $\Re(s)>0$ (stability condition)
    \item $\Re[Z(j\omega)] \geq 0, \forall \omega \in \mathbb{R}^{+}$
\end{enumerate}

The poles of (\ref{eq:impedance_siso}) are given by (\ref{eq:poles}). As the pole in the origin guarantees marginal stability, the second-order polynomial poles must have real part negative. Thus, it is possible to verify that the only requirement to assure stability is to have all gains positive, already satisfied.

\begin{equation}
    \begin{split}
        &s_1=0 \\
        &s_{2,3} = \frac{-(B_y+Pk_r)\pm \sqrt{(B_y+Pk_r)^2-4DK_yk_r}}{2Dk_r}
    \end{split}
    \label{eq:poles}
\end{equation}

\begin{figure}[t]
    \centering
    \includegraphics[width=0.9\linewidth, trim={0.0cm 0cm 0.0cm 1cm},clip]{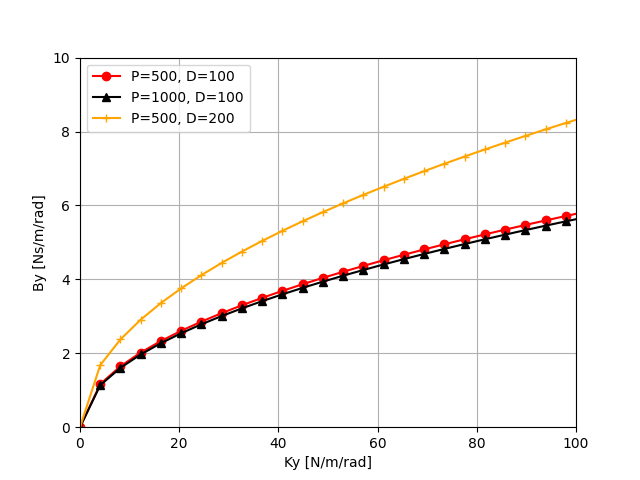}
    \caption{Theoretical Z-width for the current model based on (\ref{eq:second_criteria}). The area above the curves is the corresponding passive region. Different sets of position controller gains, $P$ and $D$, show the impact of each gain on the resulting passive range.
    }
    \label{fig:z_width}
\end{figure}

A sufficient criterion to satisfy the second passivity is demonstrated by the inequality (\ref{eq:second_criteria}).
This relation must be met to guarantee the passive behavior of the closed-loop system, which will be used in this work to set boundaries during the optimization process, as discussed in the next section. The range of passive impedances is shown in Fig. \ref{fig:z_width}.

\begin{equation}
    B_y^2D+K_y(JP-D^2k_r) > 0
    \label{eq:second_criteria}
\end{equation}

\subsection{Evolutionary optimization}

Genetic algorithms are a class of evolutionary optimization widely used for single or multiobjective tasks. They assign a combination of parameters in the search space to an individual, run the experiment, and evaluate fitness functions. They are also population-based, meaning that they repeat this process to several individuals in a generation. After all individuals of that generation are evaluated, a new generation is obtained through operations between individuals, such as mutation, crossover, tournaments, and other selection mechanisms \cite{Deb2002}. Moreover, when optimizing multiobjective problems, a set of candidates for optimal solutions are obtained as the fitnesses may have tradeoffs between them, the algorithm outcomes of the Pareto-optimal solution. Genetic algorithms can find the Pareto front for each problem, which aids the design of controllers through the experiment needs. We used the Non-Dominated Sorting Genetic Algorithm in its second version (NSGA-II) \cite{Deb2002}.

The optimization problem adopted in this work is stated in (\ref{eq:optim}): the fitness functions are represented by $f_i(\textbf{Z})$\footnote{It is known that the performance metrics are dependent on the impedance controller, the robot dynamics, and the environment. As we only can choose the controller gains, the metrics dependence on other parameters is omitted for a concise notation.}, which is the i-th value function of the $n$ objectives, and each individual $\textbf{Z}$ of the population corresponds to the parameters of admittance controller $\textbf{Z} = [B_y, K_y]^T$.
In this paper, two tradeoff metrics are chosen. First, we have decided to minimize the root mean square of the interaction torque between the robot and the subject, $f_1(\textbf{Z})=\tau_{rms}$, leading to a more transparent behavior; second, we  minimize the total amount of time needed to end the task $f_2(\textbf{Z})=T_{total}$, as it would lead to a more responsive system.

\begin{equation}\label{eq:optim}
\begin{aligned}
& \underset{\textbf{Z}}{\text{min}}
& & (f_1(\textbf{Z})=\tau_{rms}, f_2(\textbf{Z})=T_{total}) \\
& \text{s.t.} & & P, D, B_y, K_y > 0 \\
& & &  B_y^2D+K_y(JP-D^2k_r) > 0 \\
\end{aligned}
\end{equation}

The process executed  via simulation or physical experiments follows the Algorithm \ref{alg:optimization}. Constraints are evaluated after applying evolutionary optimization to avoid generating offspring outside the boundaries.

\begin{algorithm}
\SetAlgoLined
 Initialize a random population of \textit{n} individuals \\
 \For{\textbf{generation} = 1 to M}{ 
 
   \For{\textbf{individual} in \textbf{generation}}{
        Run experiment for \textbf{\textit{individual}} \\
        Evaluate the fitness of \textbf{\textit{individual}}
   }
   
   Generate \textbf{\textit{offspring}} by applying mutation, crossover, and selection
   
   \For{\textbf{individual} in \textbf{offspring}}{
        \eIf{\textbf{individual} is not passive}
        {
            Replace \textbf{\textit{individual}} with another inside the boundaries
        }
        {
            Keep \textbf{\textit{individual}}
        }
   }
   
 }
 \caption{Evolutionary optimization procedure} \label{alg:optimization}
\end{algorithm}

\section{Experiments} \label{sec:results}

\subsection{Experimental setup}

\begin{figure}
    \centering
    \includegraphics[width=0.75\linewidth, trim={0.0cm 0cm 0.0cm 0},clip]{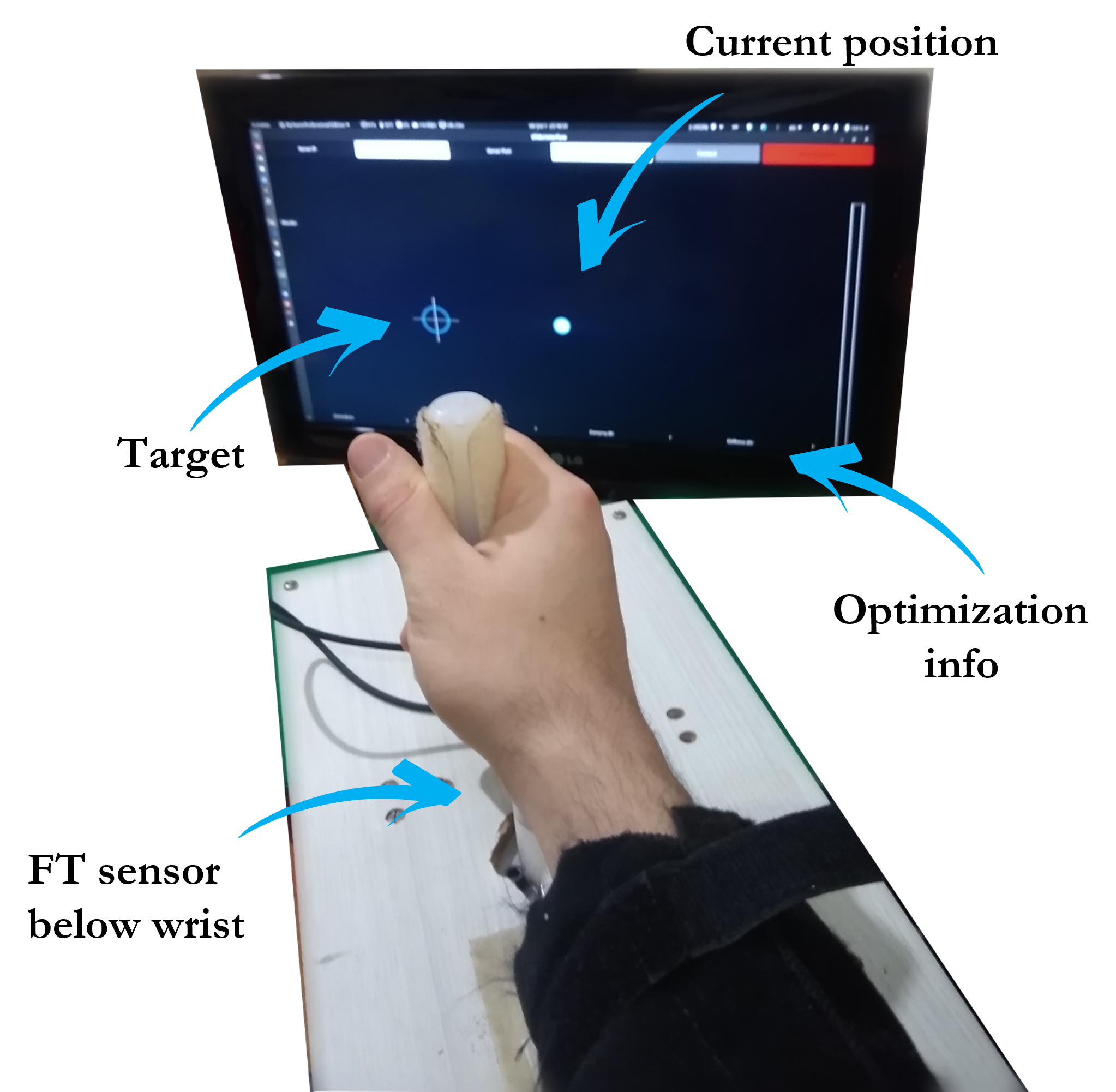}
    \caption{The experimental setup consists of a screen running the visual interface, the handler connected to the force-torque sensor, and the motor. The target has a cross shape and is activated by placing the current position marker inside it. After 2 seconds, the target moves to another position.}
    \label{fig:experimental-setup}
\end{figure}

The experimental setup (Fig. \ref{fig:experimental-setup}) contains a DC Maxon Motor RE 50, controlled via an EPOS 24/5, mounted with a gearbox of ratio $kr=3.5$. For torque feedback, a force-torque sensor model Mini40 from ATI was mounted between the handler and the motor. A BeagleBone Black runs the position and admittance controllers and communicates via CANOpen with the EPOS and the sensor.

The user is attached to the equipment and interacts with the visual interface by moving the handler, which has its current position represented on the screen by a dot marker. The user's goal is to move the current position marker within the target and keep the handler there for 2 seconds. Then, the target changes to another position. To emulate a clinical physiotherapy session for wrist rehabilitation, we aim to optimize 50 consecutive movements. Each movement consists of full flexion and full extension of the wrist, respectively. 
After executing a flexion and extension, the interaction torque is stored and used to calculate the proposed fitness metrics in (\ref{eq:optim}).

Three different subjects used the equipment, and their data were recorded. They were submitted to three experiments: a benchmark experiment without optimization, with impedance control defined by $\mathbf{Z}=[50~Nms/rad, 100~Nm/rad]^T$ executed in 5 movements; 50 movements for unconstrained optimization; and 50 more movements for constrained optimization. Each experiment had the optimization algorithm reset. We split the 50 repetitions into 10 generations of 5 individuals each, although more individuals and more generations are permissible with checkpointing throughout therapy sessions.

We conducted two analyses: the average torque profiles in time and the optimization behavior during generations regarding impedance gains and time metrics.

\subsection{Torque profiles}

\begin{figure*}
\centering
    \subfloat[]{
	\begin{minipage}[c][1\width]{0.19\linewidth}
	   \centering
	   \includegraphics[width=1\textwidth]{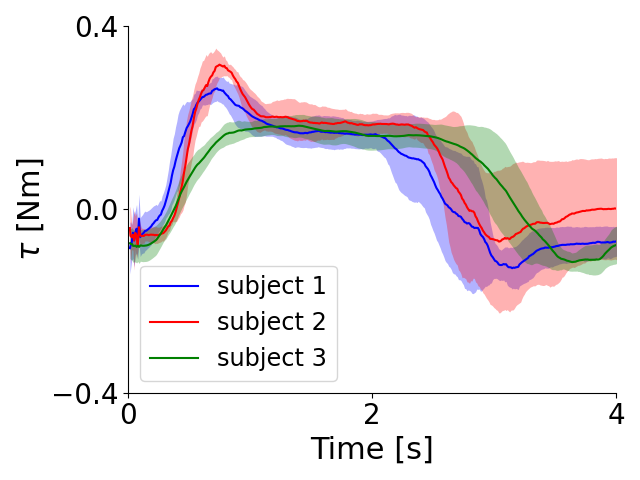}
        \label{fig:torques_no_opt}
	\end{minipage}}
    \hfill
    \subfloat[]{
	\begin{minipage}[c][1\width]{0.19\linewidth}
	   \centering
	   \includegraphics[width=1\textwidth]{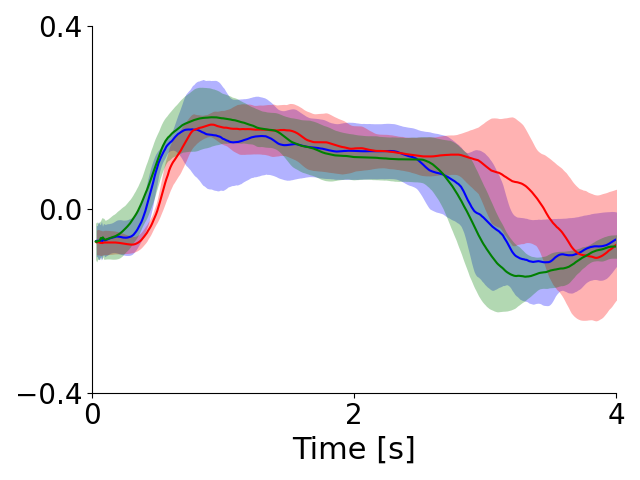}
	    \label{fig:torque_unconst_1st}
	\end{minipage}}
        \hfill
    \subfloat[]{
	\begin{minipage}[c][1\width]{0.19\linewidth}
	   \centering
	   \includegraphics[width=1\textwidth]{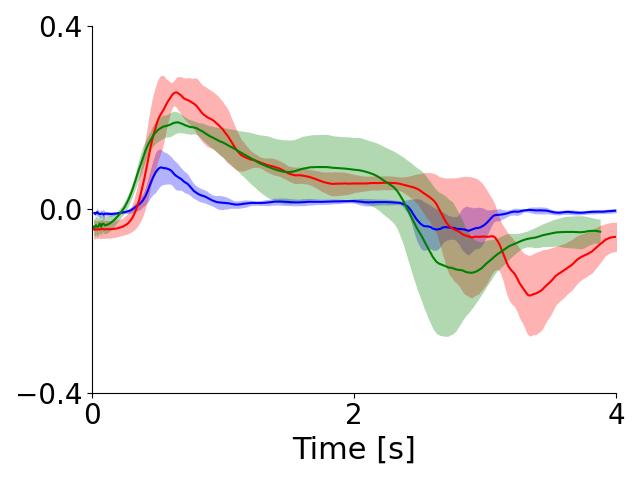}
	    \label{fig:torque_unconst_last}
	\end{minipage}}
        \hfill
    \subfloat[]{
	\begin{minipage}[c][1\width]{0.19\linewidth}
	   \centering
	   \includegraphics[width=1\textwidth]{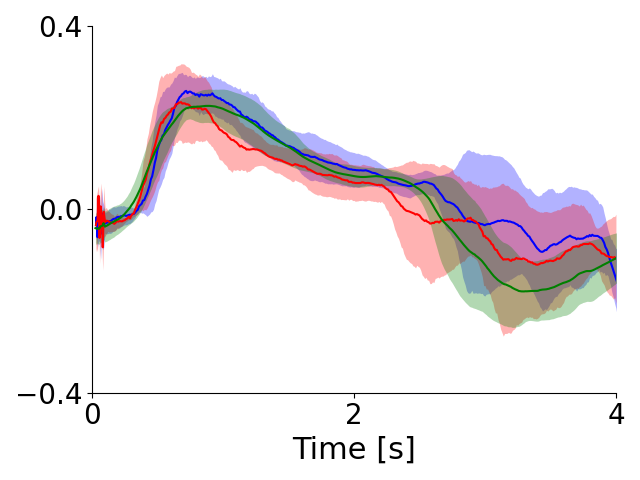}
	    \label{fig:torque_const_1st}
	\end{minipage}}
        \hfill
    \subfloat[]{
	\begin{minipage}[c][1\width]{0.19\linewidth}
	   \centering
	   \includegraphics[width=1\textwidth]{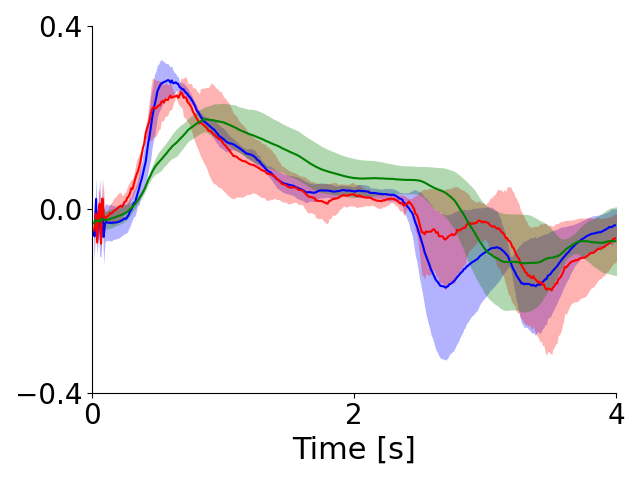}
	    \label{fig:torque_const_last}
	\end{minipage}}
	\hfill
    \caption{Mean and deviation of the torque profiles through time for all subjects considering: (a) no optimization; (b) 1st generation unconstrained; (c) 10th generation unconstrained; (d) 1st generation constrained; (e) 10th generation constrained.}
    \label{fig:torques}
\end{figure*}

Fig. \ref{fig:torques} shows the interaction torques measured by the force-torque sensor. It shows the mean and deviation for each subject on different stages for the cases of (\ref{fig:torques_no_opt}) no optimization, (\ref{fig:torque_unconst_1st}) 1st generation unconstrained, (\ref{fig:torque_unconst_last}) 1st generation unconstrained, (\ref{fig:torque_const_1st}) 10th generation constrained, and (\ref{fig:torque_const_last}) 10th generation constrained. Each subject executed each trial at different times, but all plots were clipped in 4 seconds for better visualization.

The case without optimization (\ref{fig:torques_no_opt}) has a lower deviation than both first generations with and without constraints (\ref{fig:torque_unconst_1st} and \ref{fig:torque_const_1st}, respectively), which was expected due to the lack of changes in the robot. Also, it is possible to verify that subjects 2 and 3 have similar wrist impedances due to the torque profile, while subject 1 has a significantly different impedance.

However, subjects 2 and 3 achieved different torque profiles after ten generations of optimization, as seen in \ref{fig:torque_unconst_last} and \ref{fig:torque_const_last}. This demonstrates that the evolutionary algorithm is capable of optimizing without assumptions on the environment and adapting accordingly.

We conducted a broader analysis on the performance metrics for each of the five cited experiments: all subjects were taken into account in the same mean and deviation for the torque and total time, leading to an average torque $\hat{\tau}_{rms}$ and average total time $\hat{T}_{total}$. Results are shown in Tab. \ref{tab:averages}. The case without optimization had similar torque and even a better time than the first generations for constrained and unconstrained methods. However, comparing the technique without optimization after ten generations for the other cases, it performed worse: unconstrained achieved a decrease of 40\% in torque and 6.8\% decrease in total time; for the constrained study, a 20\% decrease in torque and similar time when compared to without optimization.

\begin{table}[]
\centering
\caption{Metrics average over all subjects for the different methods studied.}
\begin{tabular}{lcc}
                        & $\bar{\tau}_{rms}$ [Nm]   & $\bar{T}_{total}$ [s] \\ \hline
No optimization         & \textbf{0.15$\pm$0.02}    & \textbf{4.86$\pm$0.57}     \\
1st gen. unconstrained  & \textbf{0.13$\pm$0.04}    & \textbf{5.48$\pm$0.91}     \\
10th gen. unconstrained & \textbf{0.09$\pm$0.04}    & \textbf{4.53$\pm$0.53}     \\
1st gen. constrained    & \textbf{0.14$\pm$0.02}    & \textbf{5.23$\pm$0.73}     \\
10th gen. constrained   & \textbf{0.12$\pm$0.02}    & \textbf{4.83$\pm$0.66}     \\ \hline
\end{tabular}
\label{tab:averages}
\end{table}

\begin{figure*}
	\begin{minipage}[c][1\width]{0.32\linewidth}
	   \centering
	   \includegraphics[width=.95\textwidth]{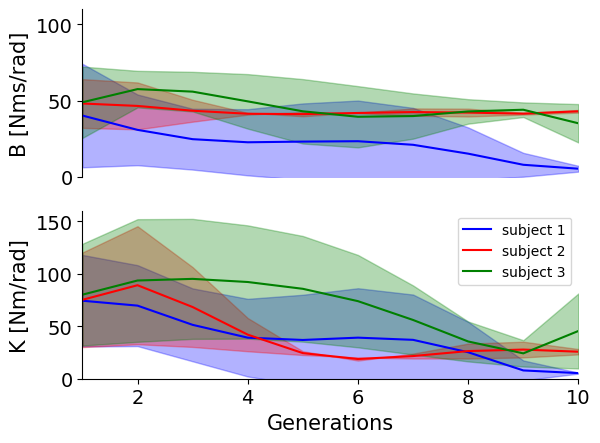}
	\end{minipage}
    \hfill
    \subfloat[]{
	\begin{minipage}[c][1\width]{0.32\linewidth}
	   \centering
	   \includegraphics[width=.95\textwidth]{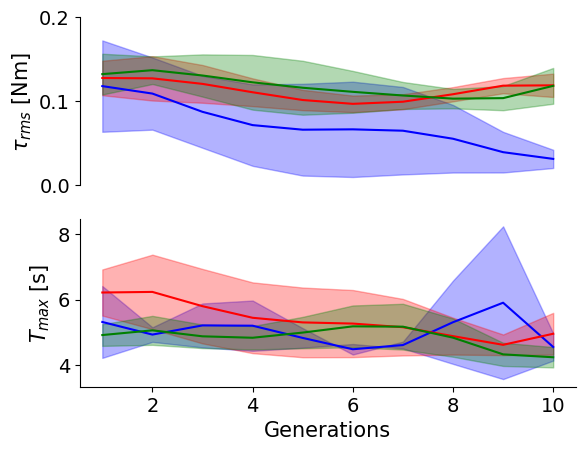}
	    \label{fig:triple_a}
	\end{minipage}}
        \hfill
	\begin{minipage}[c][1\width]{0.32\linewidth}
	   \centering
	   \includegraphics[width=.95\textwidth]{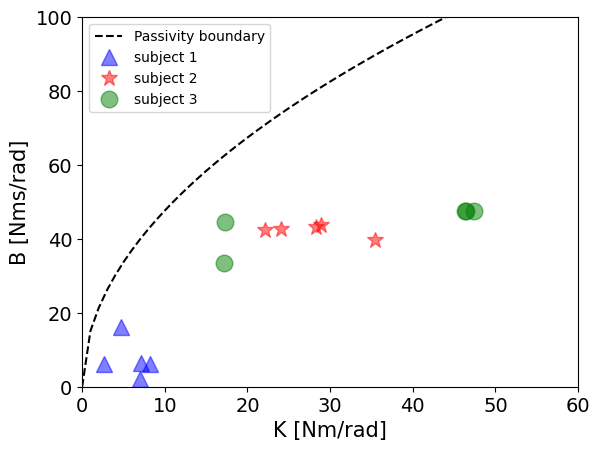}
	\end{minipage}
        \vfill
	\begin{minipage}[c][1\width]{0.32\linewidth}
	   \centering
	   \includegraphics[width=.95\textwidth]{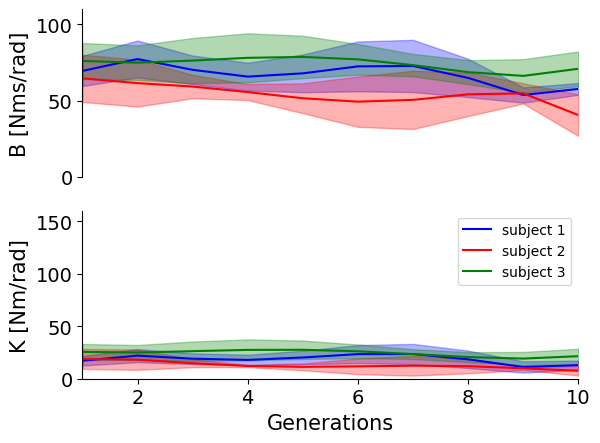}
	\end{minipage}
        \hfill
    \subfloat[]{
	\begin{minipage}[c][1\width]{0.32\linewidth}
	   \centering
	   \includegraphics[width=.95\textwidth]{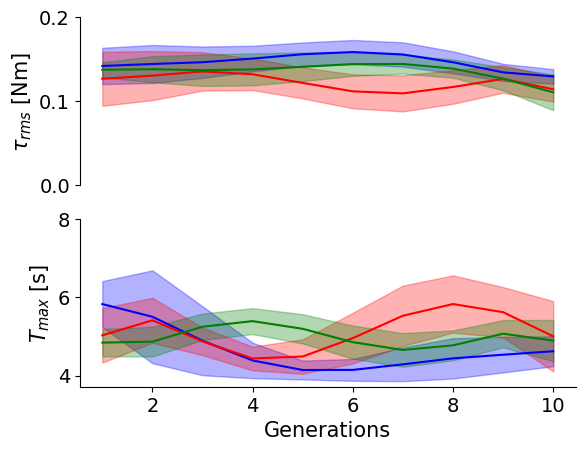}
	    \label{fig:triple_b}
	\end{minipage}}
        \hfill
	\begin{minipage}[c][1\width]{0.32\linewidth}
	   \centering
	   \includegraphics[width=.95\textwidth]{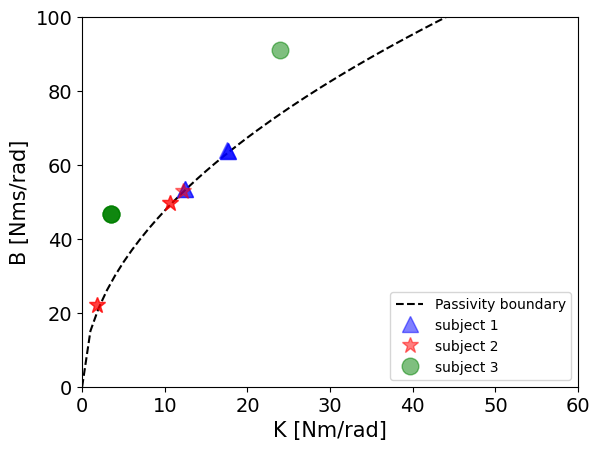}
	\end{minipage}
    \caption{The behavior of the optimization task for each generation using mean and standard deviation for all controllers in that population: (left) impedance gains, (center) metrics, and (right) last generation Z-width impedance gains. Results are shown for (a) unconstrained and (b) constrained cases.}
    \label{fig:triples_mbk_metrics_zwidth}
\end{figure*}

\subsection{Optimization over the generations}

Fig. \ref{fig:triples_mbk_metrics_zwidth} shows the optimization development for each generation for all three subjects: in the left, we have the mean and standard deviation of the impedance gains in each generation, demonstrating the distribution along with the search space; in the center, the figure is plotted the metrics' mean and standard deviation during optimization; and in the right, we can see the last generation controllers in the Z-width plot. In Fig. \ref{fig:triple_a} we have the unconstrained case and in Fig. \ref{fig:triple_b} the constrained one. 

Regarding the left plot about the controller gains, the unconstrained optimization had a high variance either for $B_y$ and $K_y$ and started to decrease after generation 8. When the optimization reached the 10th generation, the variance reduction indicated that each population was closer to the Pareto front. The average value for each generation converged to different values, illustrating the adaptive capability of the method. Meanwhile, the constrained optimization had smaller deviations since the beginning of the experiments due to reduced search space.

Concerning the optimization metrics (center plot), the optimization occurred in a similar way for all subjects in the constrained case. However, subject 1 had its interaction torque minimized for the unconstrained case but with a high deviation in the generations. The robot achieved a few unstable configurations during optimizations for subject 1, leading to some controllers with high interaction torques and higher times, as can be seen by the peak in the blue curve for the time metric. This behavior is avoided in the constrained optimization proposed.

For the Z-width mapping, the right plot shows the impedance controllers of the last generation for all subjects. The optimization, either in constrained or unconstrained setup, is trying to reach the origin of the Z-width, i.e., low stiffness and damping, which leads to smaller interaction torques. It is also possible to notice that the passivity boundaries in the optimization successfully generated passive impedance controllers for all subjects.

\section{Conclusions}
\label{sec:conclusion}

This paper presented a hybrid model-based evolutionary optimization method for improving autonomy during pHRI. The method provides an approach that allows continuous optimization as the user evolves or another user interacts with the robot. Different individuals have different arm impedances, and we use passivity theory to draw constraints on the optimization problem.

A theoretical model was developed as a proof of concept to show the method's applicability. Gains were checked through the generations, and the experiments showed the difference in behavior during the interaction of stable and passive controllers. Thus, the technique has proven independent of the environment, achieving different impedances for different subjects. Moreover, the optimization method minimized on average 20\% of the interaction torque in the case with passive boundaries, which helped maintain stability and improved performance. 

Future work considers model nonlinearities, such as sensor noise, friction, geartrain backlash, and sampling effects. Also, other metrics should be investigated to understand the optimization behavior in the process.

\bibliographystyle{IEEEtran}
\bibliography{IEEEabrv,ms.bib}

\addtolength{\textheight}{-12cm}

\section*{Acknowledgment}
Authors would like to thank AWS; CNPQ Grant 314936/2018-1; FAPESP Grants 2018/15472-9, 2017/01555-7, and 2013/07276-1; and FINEP. This study was also partially funded by the Coordenação de Aperfeiçoamento de Pessoal de Nível Superior - Brazil (CAPES) -- Finance Code 001.

\end{document}